# Operator Guidance Informed by AI-Augmented Simulations

**Samuel J. Edwards**, NSWCCD, Bethesda, Maryland, USA, samuel.j.edwards33.civ@us.navy.mil
**Michael D. Levine**, NSWCCD, Bethesda, Maryland, USA, michael.d.levine2.civ@us.navy.mil

**Abstract**

*This paper will present a multi-fidelity, data-adaptive approach with a Long Short-Term Memory (LSTM) neural network to estimate ship response statistics in bimodal, bidirectional seas. The study will employ a fast low-fidelity, volume-based tool SimpleCode and a higher-fidelity tool known as the Large Amplitude Motion Program (LAMP). SimpleCode and LAMP data were generated by common bi-modal, bi-directional sea conditions in the North Atlantic as training data. After training an LSTM network with LAMP ship motion response data, a sample route was traversed and randomly sampled historical weather was input into SimpleCode and the LSTM network, and compared against the higher fidelity results.*

## 1. Introduction

The safety of a ship and its crew in heavy weather and rough sea conditions demands proper operational guidance. Operational guidance is provided in the form of selection of speeds and headings, and is generally based on accessing ship motions response predictions from a pre-computed database or look-up table for a given condition. Operational guidance is an important consideration in the survival of a ship and has been the focus of many International Maritime Organization (IMO) publications, *IMO (1995), IMO (2007), IMO (2020)*. Recommendations for ship-specific operational guidance has been developed and discussed in the interim guidelines of the Second Generation Intact Stability by IMO, *IMO (2020)*. While these guidelines are certainly useful in design and at sea, they are not comprehensive.

The ocean environment is random and complex. Consequently a pre-computed database cannot completely capture all ocean conditions potentially encountered. Accordingly, a computationally feasible approach is needed to estimate ship responses for a range of conditions.

A simplified approach for ship motion response predictions typically assumes a unidirectional seaway with a unimodal wave spectrum. However, realistic seaways typically encompass both wind and swell components that are can be delineated in terms of wave directionality and modal frequencies. Bi-directionality and bimodal spectra are common wave characteristics that are suitable for consideration in predictive ship response models.

Multi-directionality has been considered in *Yano, et al. (2019)*, where wave radar data were invoked to generate a wave spectrum in simulations of a Ropax ship. By Grim's effective wave and a reduced-order roll equation, the maximum roll angle was estimated at various ship headings and multiple metacentric heights for the given directional wave spectrum. While the maximum roll angle is a useful metric, other seakeeping and structural response parameters are necessary for a more comprehensive investigation of extreme motions and loads.

In a recent effort, *Levine, et al. (2022),* described a data-driven model to evaluate predicted ship motions in unidirectional waves with a unimodal spectrum. In this study, data-adaptive Long Short-Term

Memory (LSTM) neural networks were investigated as part of a multi-fidelity approach incorporating Large Amplitude Program (LAMP), *Shin et al. (2003)*, and a reduced-order model known as SimpleCode. An initial assessment of this multi-fidelity approach focused on prediction of ship motion responses in waves. LSTM networks were trained and tested with LAMP simulations as a target, and SimpleCode simulations and wave time-series as inputs. LSTM networks improved the fidelity of SimpleCode seakeeping predictions relative to LAMP, while retaining the computational efficiency of a reduced-order model. The study was expanded in *Howard, et al. (2022)*, to limited combinations of bimodal, bidirectional combinations.

In this paper, this data-adaptive approach employing LAMP, SimpleCode and LSTM neural networks is evaluated for prediction of ship motions in bimodal, bidirectional seas during a simulated voyage across the North Atlantic. Simulations are performed based on the David Taylor Model Basin (DTMB) 5415, *Moelgaard (2000)*, in the most common combinations of primary and secondary wave conditions observed in the North Atlantic. Then, random observations of primary and secondary sea states along a prescribed journey are used as input into SimpleCode, LAMP, and the LSTM framework and the seakeeping statistics and time series are compared.

## 2. Methodology

### 2.1. SimpleCode and LAMP

SimpleCode is a reduced-order seakeeping simulation tool that can quickly produce reasonable results, *Smith et al. (2019)*. One of the key simplifications in SimpleCode is in the local variation of wave pressure, where the hydrostatic and Froude-Krylov equations can use volume integrals instead of integrating over the surface of the ship, *Weems and Wundrow (2013)*. With pre-computed Bonjean curves, instantaneous submerged volume and geometric center, the sectional hydrostatic and Froude-Krylov forces can be computed efficiently.

LAMP is a higher fidelity simulation tool that considers the 6-DOF forces and moments acting on the ship implemented by a 4th order Runge-Kutta solver in the time domain, *Shin et al. (2003)*. Central to the code is the solution to the 3-D wave-body interaction problem. The pertubation velocity potential is solved over the mean wetted hull surface. Hydrostatic and Froude-Krylov forces are solved over the instantaneous wetted hull surface. Within LAMP, the nonlinearities considered in the solution can be altered through how the model is represented mathematically e.g., including body non-linear hydrodynamics or large lateral motions. The version of LAMP in the current application was LAMP-3. In LAMP-3, approximate body non-linear hydrodynamics with large lateral motions are accounted for. LAMP effectively estimated motions comparable to model tests, *Lin et al. (2007)*.

LAMP is computationally intensive relative to reduced-order SimpleCode. LAMP-3 can run at nearly real-time, though some parameters such as number of wave frequency components, free surface panel definition, and hull offsets can be adjusted and increase the computational effort. For example, generation of a single realization of a 30 minute epoch entails approximately 30 minutes of compurational time. In contrast with the same number of frequency components, SimpleCode can run on the order of 5,000 independent realizations of 30 minute epoch data in 30 minutes of computational time, *Smith (2019)*.

From the perpective of fidelity, SimpleCode can produce an approximation to LAMP predictions when tuned radiation and diffraction forces are included, *Weems and Belenky (2018), Pipiras et al.*

*(2022)*. However, a fidelity gap still exists that can be potentially addressed using a data-adaptive machine learning method, *Levine et al. (2022), Howard et al. (2022)*. In this paper, this method is applied to the to bimodal and bidirectional waves.

## 2.2. Long Short-Term Memory

An LSTM network, *Hochreiter and Schmidhuber (1997)*, is a type of recurrent neural network, which incorporates both short and long-term memory based on data-adaptive learning for estimation of a function. These memory effects are stored in weight matrices, which along with other operations, transform input matrices to target output matrices. The following set of equations show the operations that occur in a LSTM layer.

$$f_1 = \sigma\left(W_{f_1} x^{[t]} + U_{f_1} h^{[t-1]} + b_{f_1}\right) \tag{1}$$

$$f_2 = \sigma\left(W_{f_2} x^{[t]} + U_{f_2} h^{[t-1]} + b_{f_2}\right) \tag{2}$$

$$f_3 = tanh\left(W_{f_3} x^{[t]} + U_{f_3} h^{[t-1]} + b_{f_3}\right) \tag{3}$$

$$f_4 = \sigma\left(W_{f_4} x^{[t]} + U_{f_4} h^{[t-1]} + b_{f_4}\right) \tag{4}$$

$$c^{[t]} = f_1 \odot c^{[t-1]} + f_2 \odot f_3 \tag{5}$$

$$h^{[t]} = f_4 \odot \tanh\left(c^{[t]}\right) \tag{6}$$

Here, $W$ and $U$ are weight matrixes, $b$ are the bias vectors, $x^{[t]}$ is the input vector at time $t$, $h^{[t]}$ is the hidden state vector at time $t$, $c^{[t]}$ is the cell state vector at time $t$, $\sigma$ is the sigmoid function, *tanh()* is the hyperbolic tangent function, and $\odot$ represents the Hadamard product. The output or target at time $t$ is equal to the hidden state vector at time $t$, $h^{[t]}$. The weight matrices and bias vectors are progressively adapted during the training process to minimize the specified loss between the training data and test data. Mean-squared error is the loss function to quantify the error between the training and test sets. The formula for the mean-squared error (MSE) is in the following equation, where

$$MSE = \frac{1}{N} \sum_{i=1}^{N} \left(y_T(t_i) - y_L(t_i)\right)^2 \tag{7}$$

Here, $N$ is the number of points in the time series; $y$ is the response matrix of the time series for heave, roll, and pitch; subscript $T$ is the target time series, subscript $L$ is the LSTM produced time series; and $t_i$ is the *i-th* time instant in the time series.

The current framework encompassed three consecutive LSTM layers followed by a dense layer to transform the input into the target output. LSTM inputs were heave, roll, and pitch responses provided from SimpleCode as well as the wave elevation at the center of gravity of the ship at the ordered speed and heading and the corresponding slope of the wave field in the x- and y-directions to account for the bi-directional seas. The target time series were the heave, roll, and pitch motions predicted by LAMP in 6-DOF simulations. For higher fidelity accurate prediction of these vertical degrees of freedom (heave, roll and pitch), the horizontal degrees of freedom (surge, sway, and yaw) were included in the LAMP simulations. Subsequent training of the LSTM network specifically focused on the relevant target outputs for the heave, roll and pitch motions. A series of LSTM networks are trained on a given set bimodal data and then tested on different bimodal systems sampled across the North Atlantic voyage.

**2.3. Experimental Set-up**
Simulations were performed with the DTMB 5415, *Moelgaard (2000)*. A rendering of the Model 5415 is in Figure 1.

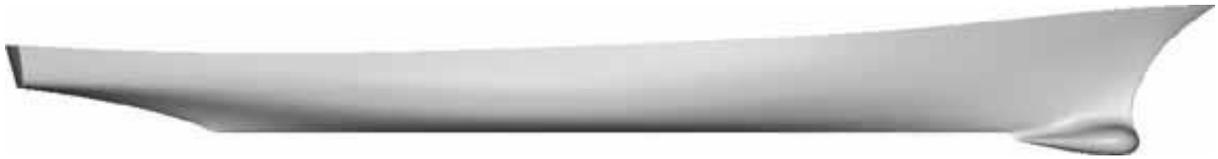

Fig.1: Rendering of the DTMB 5415.

The basic parameters of the full-scale DTMB 5415 are in Table I.

Table I: Primary parameters of the full-scale DTMB 5415.

| Parameter | Value |
| --- | --- |
| Lwl [m] | 142.0 |
| B [m] | 19.06 |
| T [m] | 6.51 |
| Displacement [t] | 9,156.38 |
| KG, from BL [m] | 7.71 |
| LCG, from FP [m] | 72.1 |

To determine the scope of training data environmental conditions, the 100 most historically observed combinations of primary and secondary seas in the North Atlantic during the month of December were used. The historical records were sourced from Wavewatch III hindcasts run by the US National Oceanic and Atmospheric Administration, *Tolman (2009)*. The area defined as the North Atlantic is in Figure 2.

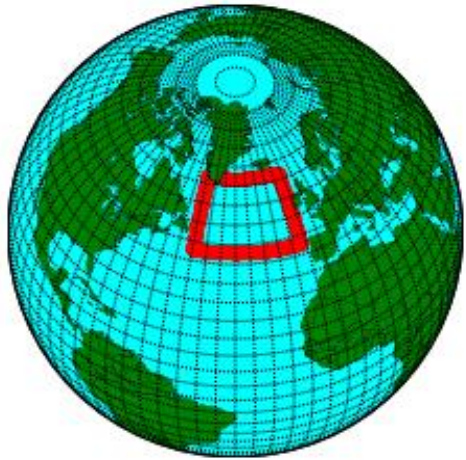

Fig.2: Bounds considered as "North Atlantic" to sample historical wave parameters.
For this study, primary spectra characterizing wind-generated waves and secondary spectra characterizing swell were formed from ITTC (International Tow Tank Conference) spectra, *ITTC (2014)*. In addition, a ship speed of 10.0 kts was considered along with primary relative wave headings from 0 to 330 degrees in 30 degree increments. In this paper, 0 degrees is defined as head seas and 180 degrees is following seas. Secondary spectra wave directions were determined based on the most probable difference between the primary and secondary sea directions.

SimpleCode was set to run in the 3-DOF (heave, roll, and pitch,) configuration while LAMP was configured to run in 6-DOF with a Proportional-Integral-Differential (PID) rudder controller. The difference in configurations between LAMP and SimpleCode can result in different global positions within respective runs due to SimpleCode being restricted in sway and yaw while the controller in LAMP attempts to keep the ordered heading but ultimately includes variations in position. As a result, the simulated ships experience different wave elevations and forces at the center of gravity. The distinction in experienced waves consequently causes phase shifts between the SimpleCode and LAMP time series that may increase as the simulation progresses and affect the LSTM performance. A 6-DOF version of SimpleCode with a similar PID controller to LAMP would mitigate the difference in experienced waves and likely improve performance. In the current structure, each realization was 1,920 seconds (including an initial 120 second wave ramp-up) with a time step length of 0.05 seconds. In total, 5 realizations were generated for each of the 2,400 combinations of conditions in both SimpleCode and LAMP for training, validation, and testing.

LSTM networks were then trained based on simulations of the most observed North Atlantic in December bi-modal sea states in SimpleCode and LAMP. A network was trained for each of the 12 primary headings. From the 12,000 runs, 50 were randomly selected from each primary relative wave heading for training for each network, 25 were randomly selected for validation, and 25 were selected for testing. Table II details the hyperparameters in the training process.

Table II. Defining hyperparameters of the LSTM framework.

| Hyperparameter | Value |
| --- | --- |
| Time steps, N | 18,000 |
| Time resolution factor | 9 |
| Hidden state size | 150 |
| Number of LSTM layers | 3 |

Once the LSTM network was trained, a Great Circle route between Norfolk, USA and Bergen, Norway was generated. Random samples of primary and secondary seas based on the same historical wave database histogram during the month of December were generated along the Great Circle path. The simulation length for both LAMP and SimpleCode were the same as in the training stage. For each observation along the generated path, five simulations were run in SimpleCode and LAMP for a constant ship speed of 10 knots. The SimpleCode time series were then standardized by the training data statistics run through the trained LSTM network. Histograms of the standard deviation of heave, roll, and pitch along the path for SimpleCode, the LSTM-corrected time series, and LAMP were compared. Additionally, the time series from SimpleCode, the LSTM-correction, and LAMP in the sea state with the largest significant wave height were compared.

**3. Results**

## 3.1. LSTM Training

As mentioned in Section 2.3, the training and validation realizations for each network were randomly selected from combinations of the 100 most common combinations of primary and secondary significant wave height, primary and secondary modal period, and difference in direction between primary and secondary seas. The training success varied for each of the networks, which were partitioned by primary relative wave direction, as a result of variation in secondary sea heading. Prior to training, the SimpleCode input and LAMP target realizations were standardized by the training data statistics e.g., heave, roll, and pitch standard deviations and means. The training and validation error plot for the primary relative wave direction of 330 degrees over 100 training epochs is in Figure 3.

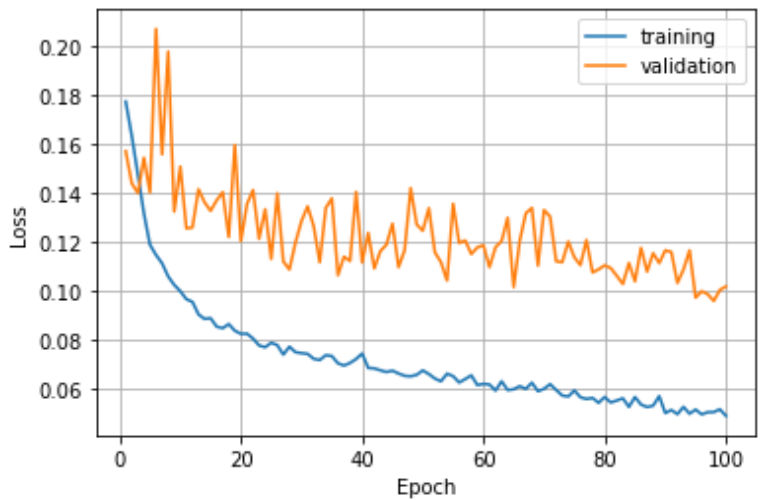

Fig.3: Training and validation error for a relative primary wave direction of 330 degrees over 100 epochs.

The loss in Figure 3 is the total mean square error between the standardized LSTM output and LAMP for heave, roll, and pitch.

The training and validation set sizes were limited by a single 4-GB Graphical Processing Unit (GPU) and by the length (19,200 points) of each realization. This limitation was the primary reason for randomly sampling realizations with different parameter combinations so that bias could be reduced without losing accuracy. Still, the not including more combinations of primary and secondary relative wave directions through increase in training and validation set sizes did affect overall performance – especially in cases where parameter combinations varied significantly from the training data set. However, the LSTM still provided an improvement in estimation of statistics compared to SimpleCode.

## 3.2. North Atlantic Journey Statistical Comparison

The primary goal of this study was to generate statistics for pertinent seakeeping motions in bimodal, bidirectional seaways common in the winter in the North Atlantic for operational guidance. An example journey was plotted from Norfolk, USA to Bergen, Norway. The route was set through half-degree latitude and longitude grids for which weather data had been collected and was set to be the shortest distance possible, or a Great Circle route. The route is in Figure 4.

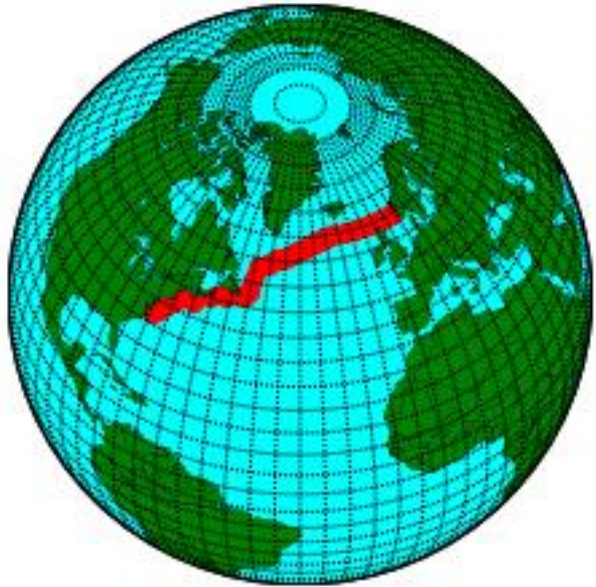

Fig.4: Adjusted great circle route between Norfolk, USA and Bergen, Norway.

Over the route, primary and secondary wave parameters were randomly selected from historical weather observations from each latitude-longitude grid and the corresponding conditions were run in SimpleCode, LAMP, and through the trained LSTM networks. Although, the random selection of observations does not model the inherent dependence between wave parameters of adjacent/nearby grids, the imposed variation here provides a reasonable initial test of the LSTM networks. The standard deviation from each simulation was estimated over five, 30-minute realizations. The standard deviations from heave, roll, and pitch were tabulated, binned, and counted over the example voyage to provide a summary of the journey. Kernel density estimations of the standard deviation probability density functions (pdfs) for each considered degree of freedom are in Figures 5-7.

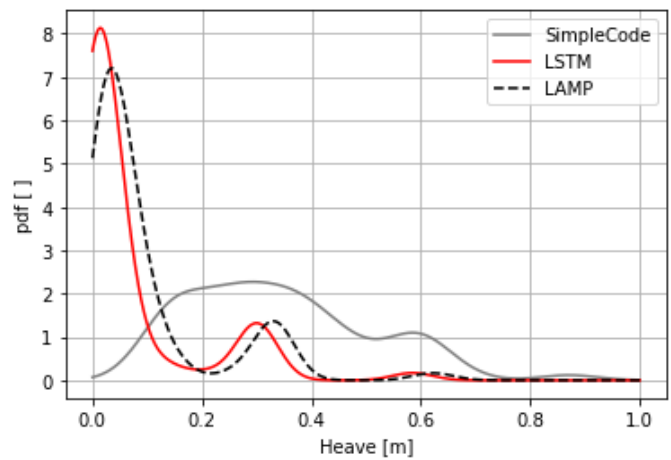

Fig.5: Heave standard deviation pdfs from SimpleCode, the LSTM networks, and LAMP.

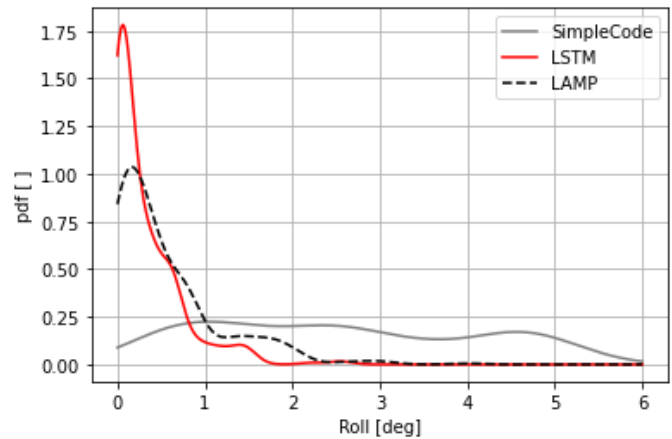
Fig.6: Roll standard deviation pdfs from SimpleCode, the LSTM networks, and LAMP.

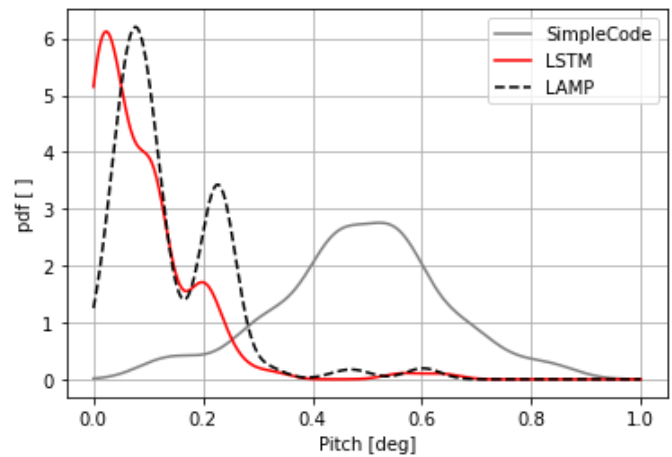
Fig.7: Pitch standard deviation pdfs from SimpleCode, the LSTM networks, and LAMP.

In each degree of freedom, SimpleCode generally over-predicts LAMP while the LSTM networks improve the estimation while generally under-estimating LAMP. In SimpleCode, the larger, more varied responses are likely due to the hydrodynamic forces being concentrated on strictly the presented 3-degrees of freedom. For example, all lateral forces acting on the hull in SimpleCode can only go into producing roll; when in reality, some of the produced movement goes into sway or even yaw. But since SimpleCode is constrained in the lateral frame in oblique seas, roll is over-estimated and varies widely. The LSTM improves upon SimpleCode in roll by not only moving the most probable standard deviation closer to the most probable LAMP standard deviation, but also in reducing the variation in standard deviation. The estimation of heave and pitch by SimpleCode is generally more accurate but is still larger than the LAMP estimates. Again, the LSTM mostly captures the peak behaviour seen in the LAMP observations and reduction in variation compared to SimpleCode but is generally leads to under-prediction.

In Figures 5-7, the LSTM provided an improvement compared to SimpleCode with respect to LAMP in capturing the seakeeping summary. To further test the LSTM, the results from the worst conditions, defined as largest primary and secondary significant wave heights, resulting from the random selection of conditions over the journey were compared to LAMP and SimpleCode. In Figures 8-10, the time series from heave, roll, and pitch from SimpleCode, the LSTM networks, and LAMP are compared for a primary significant wave height of 3.0 meters, a primary wave period of 6.5 seconds, a primary wave direction of 240 degrees, a secondary wave height of 1.5 meters, a secondary wave period of 11.5 seconds, and a secondary wave direction of 330 degrees.

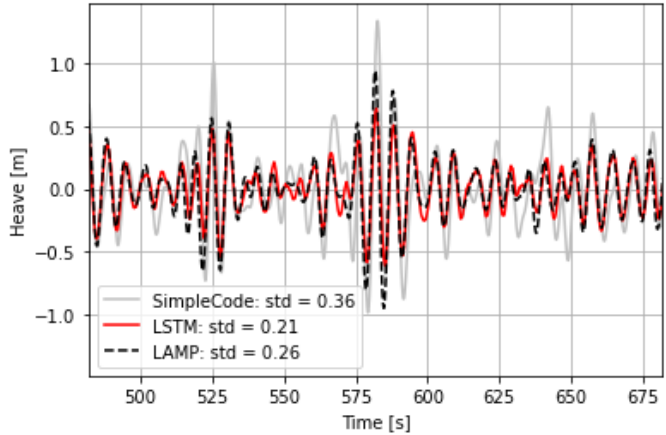

Fig.8: Heave time series snippet centered about the largest LAMP heave response.

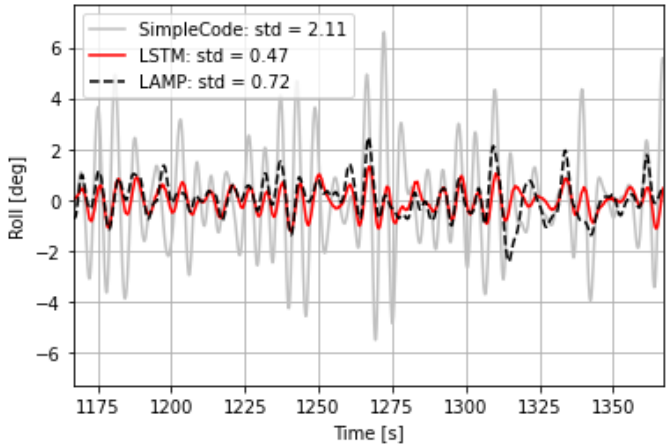

Fig.9: Roll time series snippet centered about the largest LAMP roll response.

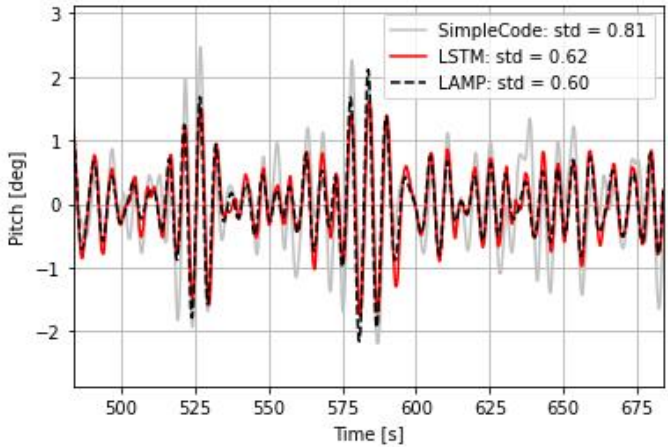

Fig.10: Pitch time series snippet centered about the largest pitch heave response.

In Figures 8-10, the LSTM improves in level of response and phasing similarities with LAMP

compared to SimpleCode. The LSTM predicted standard deviations from these selected time series compared to LAMP improved with respect to SimpleCode. In roll, the absolute percentage error of the LSTM estimated standard deviation compared to LAMP was 34.7% but it is a significant improvement to the 193% error in the SimpleCode roll standard deviation. The LSTM also generally captured the LAMP peak, which is centered in each time series snippet, in each degree of freedom in both location and magnitude.

## 4. Conclusion

Accurate predictions of the ship motion statistics are vital for operational guidance. When considering bimodal, bidirectional sea spectra, estimates of the ship response are not simple. Furthermore, generating response statistic lookup tables as functions of the many combinations of parameters sourced from high-fidelity simulations is computationally prohibitive. A data-driven approach to capture the high-fidelity response while taking advantage of lower-fidelity tools can provide a potential answer.

In this paper, LSTM neural networks were trained to improve low-fidelity, 3-DOF hydrodynamic simulations in bimodal, bidirectional seas run by SimpleCode to estimate the 3-DOF seakeeping motions of interest sourced from higher-fidelity 6-DOF hydrodynamic simulations in the same bimodal, bidirectional seas run in LAMP. Networks were trained with the most common combinations of wave parameters in the North Atlantic in December. The networks provided improved estimates of the statistics relative to LAMP compared to SimpleCode. Incorporation of the inherent dependence between the wave parameters of adjacent/nearby grids could help demonstrate the utility of fast SimpleCode-LSTM seakeeping predictions along realistic routes for path planning.

While the current LSTM framework improved upon SimpleCode, the flexibility and accuracy could be increased by expanding the size of the training and validation sets. Running the training on a larger/multiple GPUs or a more optimized set-up would allow for more variation in training and validation data. These more comprehensive training and validation sets would not only improve the network by increasing the amount of data but also general flexibility.

Additionally, a future 6-DOF SimpleCode could enhance the performance of the LSTM. The fuller accounting of force distribution could potentially allow the LSTM to focus on other improvements to reduced-order SimpleCode predictions.

## 5. Acknowledgements

The work described in this paper has been partially funded by the Office of Naval Research (ONR) under Dr. Woei-Min Lin. The work has also been funded by the Department of Defense SMART SEED Grant. The authors would also like to thank Dr. Kenneth Weems for assistance with SimpleCode and LAMP.